\documentclass[10pt,twocolumn,letterpaper]{article}

\usepackage{url}
\usepackage{kotex}
\usepackage{cvpr}
\usepackage{times}
\usepackage{epsfig}
\usepackage{graphicx}
\usepackage{amsmath}
\usepackage{amssymb}
\usepackage{subcaption}
\usepackage{multirow}
\usepackage{hhline}

\DeclareMathOperator*{\argmax}{argmax}
\DeclareMathOperator*{\argmin}{argmin}

\usepackage{caption}
\usepackage[labelformat = empty,position=top]{subcaption}

\newif\ifdraft\drafttrue
\newcommand{\specialcell}[2][c]{%
  \begin{tabular}
  [#1]{@{}c@{}}#2
  \end{tabular}}

\usepackage[breaklinks=true,bookmarks=false]{hyperref}

\cvprfinalcopy % *** Uncomment this line for the final submission

\def\cvprPaperID{2914} % *** Enter the CVPR Paper ID here

\ifcvprfinal\fi

\begin{document}

%%%%%%%%% TITLE
\title{Image Restoration by Estimating Frequency Distribution of Local Patches}

\author{
Jaeyoung Yoo$^{1,2,*}$ 
\qquad \qquad
Sang-ho Lee$^{1,*}$
\qquad \qquad 
Nojun Kwak$^{1}$ \\
$^{1}$Seoul National University\\
$^{2}$NALBI Inc.\\
% Gwanak-ro 1, Gwanak-gu, Seoul, Republic of Korea\\
{\tt\small \{yoojy31|shlee223|nojunk\}@snu.ac.kr}
% {\tt\small yoojy31@nalbi.ai, shlee223@snu.ac.kr, nojunk@snu.ac.kr}
%\and
%Sang ho lee\\
%Institution2 address\\
%{\tt\small shlee223@snu.ac.kr}
%\and
%Nojun Kwak\\
%Institution3\\
%Institution3 address\\
%{\tt\small nojunk@snu.ac.kr}
}

\maketitle
%\thispagestyle{empty}

%%%%%%%%% ABSTRACT
\begin{abstract}
In this paper, we propose a method to solve the image restoration problem, which tries to restore the details of a corrupted image, especially due to the loss caused by JPEG compression.
We have treated an image in the frequency domain to explicitly restore the frequency components lost during image compression. In doing so, the distribution in the frequency domain is learned using the cross entropy loss. 
Unlike recent approaches, we have reconstructed the details of an image without using the scheme of adversarial training. Rather, the image restoration problem is treated as a classification problem to determine the frequency coefficient for each frequency band in an image patch. In this paper, we show that the proposed method effectively restores a JPEG-compressed image with more detailed high frequency components, making the restored image more vivid. 
   
\end{abstract}

\section{Introduction}
{\let\thefootnote\relax\footnotetext{{
$^{*}$J. Yoo and S. Lee equally contributed to this work. \\ 
%This work was done when J. Yoo was with Nalbi company. \\
This work was supported by NRF of Korea (2017M3C4A7077582) and NALBI Inc.}}}

%\footnotetext{J. Yoo and S. Lee equally contributed. \\
%This work was done when J. Yoo was with Nalbi company. \\
%This research was supported by Next-Generation Information Computing Development Program through the National Research Foundation of Korea (NRF) funded by the Ministry of Science and ICT (2017M3C4A7077582).
%}

As multimedia and the Internet have become indispensable in our ordinary life, low quality compressed images are used more often because the quality of images and the consumption of data resource are highly correlated. 
In this environment, tasks such as compression artifact removal that removes artifact from a lossy-compressed image and restoration of a high quality image have recently become important areas of computer vision.

\begin{figure} 
 	\centering
	\begin{center}
	\begin{subfigure}{0.47\textwidth} % width of left subfigure
		\includegraphics[width=\textwidth]{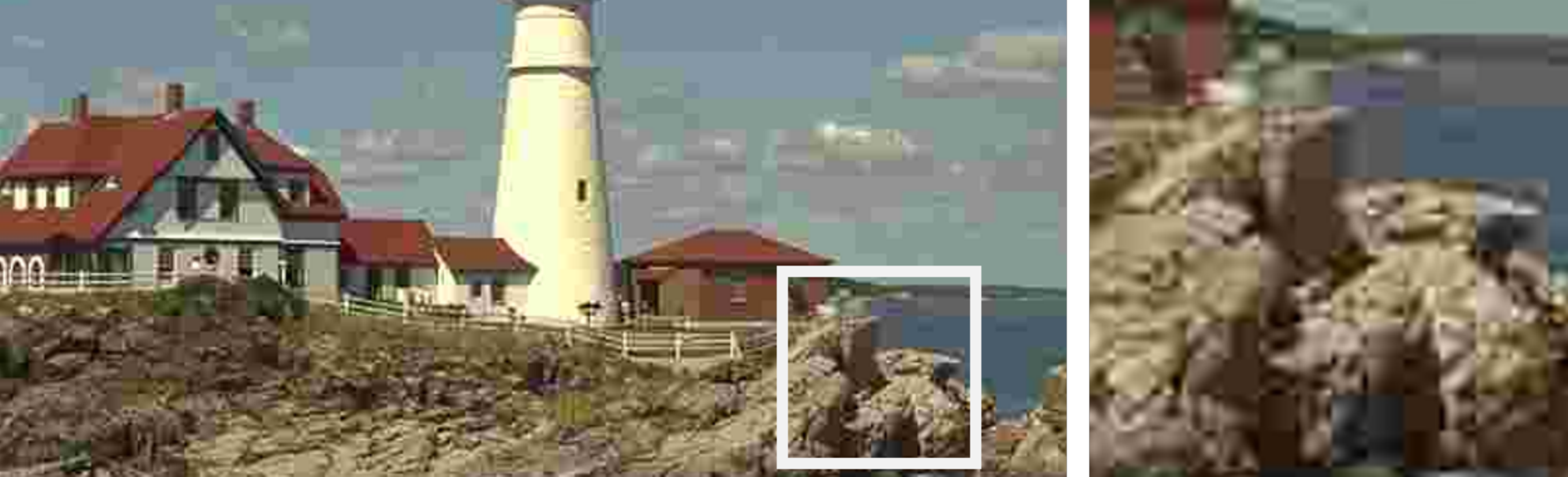}
	\end{subfigure}
    
 	\vspace{1mm} % here you can insert horizontal or vertical space
	\begin{subfigure}{0.47\textwidth} % width of right subfigure
		\includegraphics[width=\textwidth]{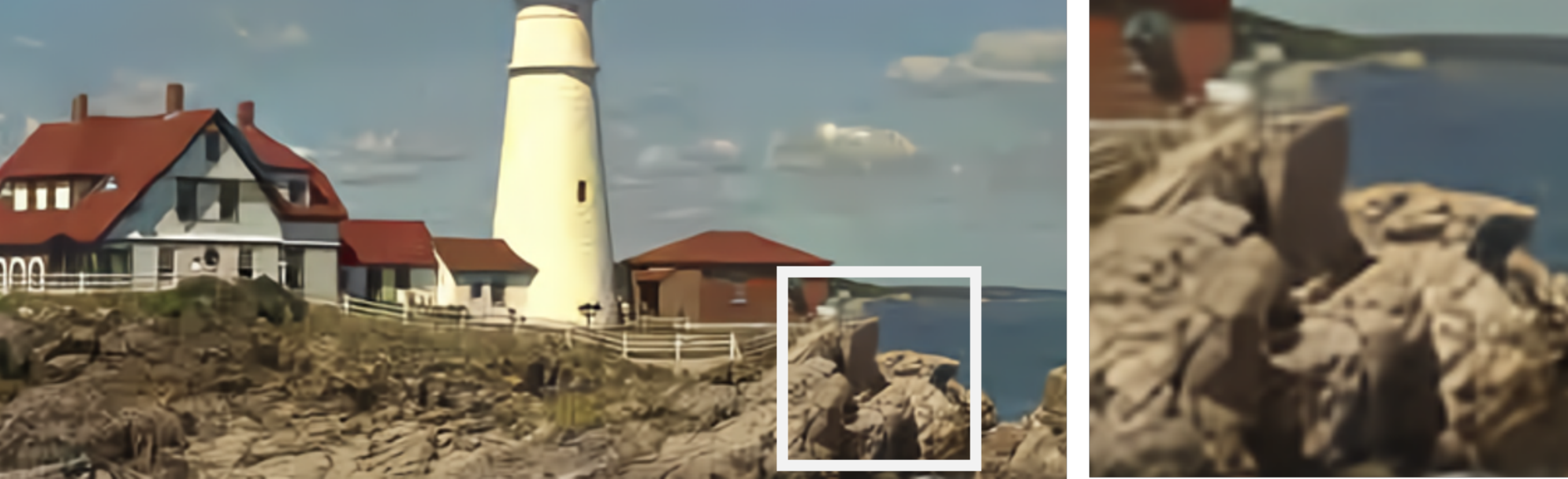}
	\end{subfigure}
    
 	\vspace{1mm} % here you can insert horizontal or vertical space
	\begin{subfigure}{0.47\textwidth} % width of right subfigure
		\includegraphics[width=\textwidth]{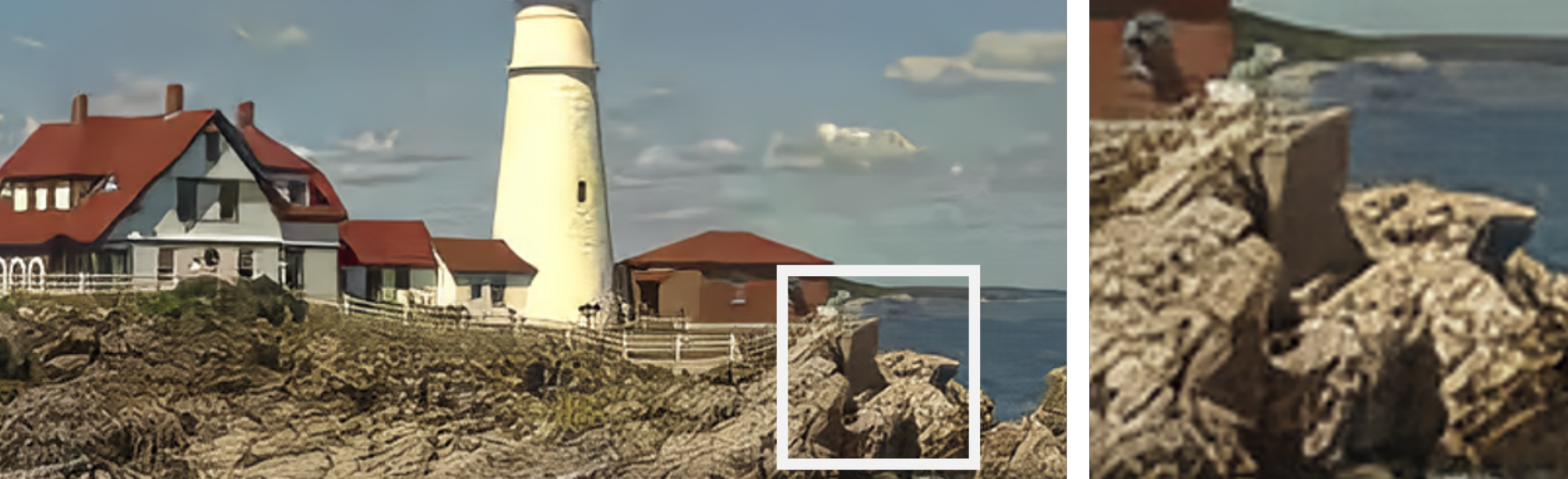}
	\end{subfigure}    
    \end{center}
	\caption{From top to bottom, the first image is a JPEG compressed image with quality factor 10, the second is a restored result by our baseline method that uses a typical encoder-decoder model whose loss function is pixel-wise MSE.
    The third is a restored result of our method, which utilizes the classifier network, trained to predict frequency distribution of possible outputs. Our result has more detail than that of the baseline.}   
    \label{fig:com1}
\end{figure}

Among various image formats, JPEG is the most commonly used lossy image compression format. 
A lossy compression like this reduces the volume of data by permanently removing some of its information. 
Thus, the reverse process of restoration of an image is basically a problem of generating information that the input image does not have, and is therefore an ill-posed problem. 
In most cases, there are several possible output images corresponding to a given input image and the problem can be seen as a task of selecting the most proper one from all the possible outputs. 
That is, the image restoration problem can be formulated as the problem of estimating the distribution conditioned on the input image.

Since convolutional neural networks (CNNs) have been actively applied in the field of computer vision, there have been many attempts to recover the lost information of an image due to compression, using CNN. These methods mainly approximate the mapping function from the input image to the output image using CNNs in a supervised manner. 
In most cases, the mean squared error (MSE) or the mean absolute error (MAE) between the output and the target image is minimized.
These approaches generally achieve good performances in commonly used metrics like PSNR (peak signal to noise ratio) and SSIM (structural similarity)~\cite{ssim}, but the output images are blurry for human eyes with absence of high frequency details as can be seen 
in the second image of Figure \ref{fig:com1}.
The reason can be attributed to the fact that the learning method tries to minimize a loss function like MSE and MAE, which is based on the pixel distance between the output and the target, forcing the model to converge to the mean or the median of all the possible solutions~\cite{arcnn,pavelsvoboda,wang2016d3,cavigelli2017cas}.

In recent years, many studies have shown good results by applying generative adversarial networks (GAN)~\cite{goodfellow2014generative} to the problems of lossy compression artifact removal~\cite{onetomany, l.galteri2017gan}. However, since training a GAN actually requires to find a Nash equilibrium between the generator and the discriminator, the learning is unstable and difficult, thus still it is very hard to reproduce the good results reported in the original papers~\cite {goodfellow2016nips}.

In this work, instead of using a generative model, we treat the image restoration problem as a classification task.
The frequency distribution of the target image is directly estimated from the input image using the cross-entropy loss function. 
By using this information together with the existing encoder-decoder neural network model, the output image can be brought closer to a natural image.
Our method, as shown in the third row of Figure \ref{fig:com1}, can generate more sharp output images with realistic details.
The contribution of this paper is threefold: 

\begin{enumerate}
\item The image restoration problem is reformulated as a task of recovering original frequency components, thus the viewpoint of an image is changed from the pixel domain to the frequency domain to explicitly recover the lost high frequency information.

\item Unlike previous works that tackle the image restoration problem by solving a regression problem or by using a generative model, we estimated the distribution of the lost information of an image by treating this task as a classification problem in the frequency domain.

\item Especially, the proposed method is applied to the tasks of JPEG compression artifact removal
and the results show that our work restores high frequency components well and produces visually satisfactory outputs.

\end{enumerate}

%----------------------------------------------------------------------------
\section{Related works}
\label{sec:rel}
Since deep neural networks (DNN) have attracted researchers' interests, many studies have been conducted to remove the lossy compression artifacts with DNN, most of which focused on enhancing the quality of JPEG images. 
Many have attempted to solve this problem by forwarding a lossy compressed image to a DNN to obtain a restored image directly from the output. The work in \cite{arcnn} is an early study of applying the DNN to the artifact removal, in which a relatively light neural network was used with pairs of a lossy compressed image and the corresponding lossless image. Better results were obtained in \cite{pavelsvoboda} by adding a loss function that emphasizes the edges using the Sobel filter.

Some attempts have been made to remove artifacts from lossy compression using discrete cosine transform (DCT), which is highly utilized in image compression algorithms such as JPEG and MJPEG.
% DCT is a method to convert an image into a frequency domain with a cosine basis function. 
In lossy compression, it is used to remove high frequency components with low energy using the fact that frequency components are well separated by frequency bands.
In the case of \cite{liu2015data, guo2016building, wang2016d3}, the compression artifact removal problem was approached from both pixel domain and DCT domain using a neural network.

Most works in this line of research cast the image restoration problem in the framework of regression and tried to minimize the loss function defined as a distance in the pixel domain, which has the disadvantage that the resultant images are blurred because the neural network takes pixel-wise average~\cite{srgan,attgan}. Some studies have attempted to solve this problem by using a classification framework directly or indirectly. Zhang \etal \cite{zhang2016colorful} showed good performances in colorization by directly classifying color pixels from grayscale pixels. Iizuka \etal \cite{iizuka2016lettherebecolor} also dealt with colorization and mitigated the disadvantages of minimizing distance loss by mixing high level features learned in classification as prior knowledge.

\begin{figure*} [t] %%%
\begin{center}
\includegraphics[width=0.9\linewidth]{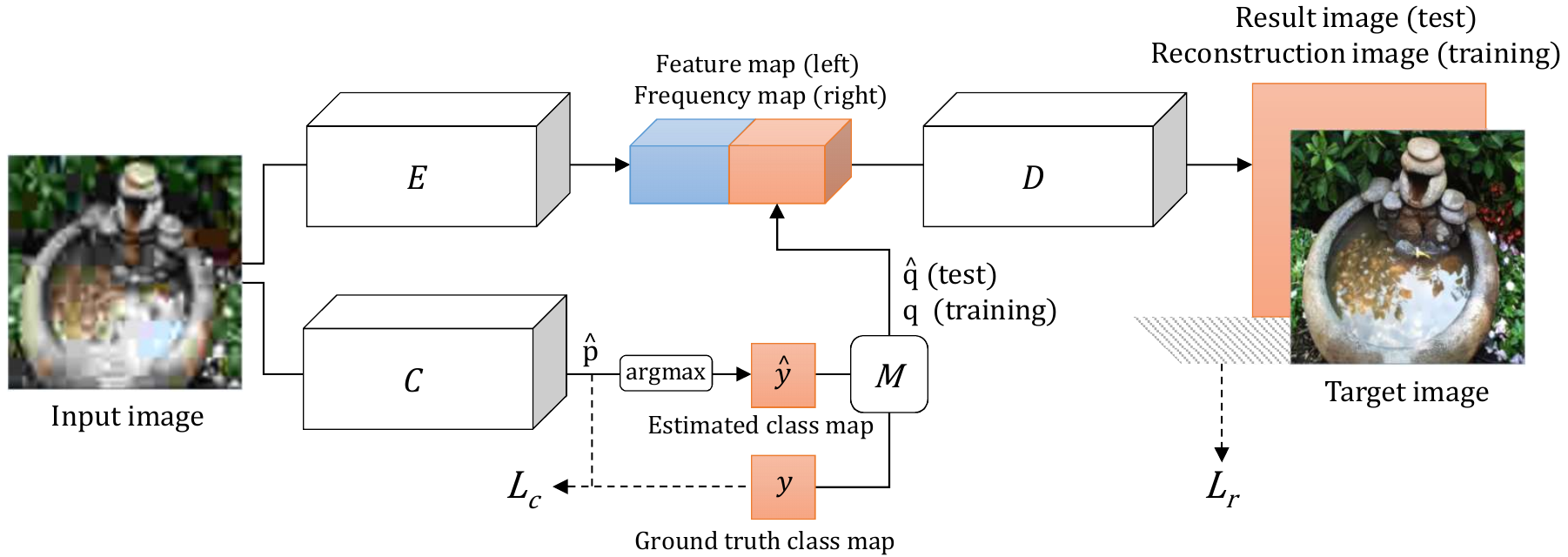}
\end{center}
\caption{\textbf{The overview of the proposed image restoration process.} The proposed network is composed of an encoder ($E$), a decoder ($D$) and a classifier ($C$).
$E$ outputs a feature map (blue) which are considered to have low frequency information, while $C$ outputs estimated frequency components for each image patch (orange). Taking the concatenation of the two as an input, $D$ outputs a reconstructed image. 
The solid line represents the forward path and the dashed line represents the loss calculation path.
}
\label{fig:ourmethod}
\end{figure*} %%% 

GAN~\cite {goodfellow2014generative} is another way to solve this problem of artifact removal and image quality enhancement.
% It is a method of adversarially training the two networks: a generator and a discriminator. 
The work \cite{onetomany} applied a GAN to remove lossy compression artifacts, which obtained better results using the GAN loss, in combination with the perceptual loss \cite{johnson2016perceptual} and JPEG-related loss. Galteri \etal \cite {l.galteri2017gan} also tackled the problem of artifact removal using the GAN. Ledig \etal \cite{srgan} suggested SRGAN, a way to create a super resolution image using a GAN, and showed that the GAN generates more realistic images compared to the conventional methods.
However, training a GAN is very difficult and unstable. 
Mescheder \etal \cite{mescheder2017numerics} pinpoints the difficulties of convergence of GAN with the recent training algorithms and proposed a new methods.
% Mescheder \etal \cite{mescheder2017numerics} numerically explained that it is difficult to train a GAN with the recent training algorithms and proposed a new training algorithm.
% Salimans \etal \cite{salimans2016improved} suggested techniques for helping training of GAN.

Unlike the previous studies which apply GAN to solve the problems caused by dealing with the image restoration as a regression problem in the pixel domain, we utilize not only spatial representations but also frequency representations for the problem. More specifically, a classification method is used for estimating the frequency distribution, which is further utilized in a regressional framework whose loss function is based on pixel-wise distance. Also, we propose a new architecture and training scheme to do this efficiently.

%------------------------------------------------------------------------
\section{Estimating Frequency Distribution for Image Restoration}
\label{sec:ourmethod}
\subsection{Problem formulation}

The neural network approaches to the existing artifact removal problem mainly try to minimize the pixel-wise distance between the ground truth $I^G$ and the restored output $I^R$ for the JPEG-compressed input $I^J$ in the sense of MSE \cite{arcnn,pavelsvoboda} or MAE \cite{lim2017enhanced}. The output images obtained with this distance-based loss are good in the sense of MSE-based metrics such as PSNR and SSIM, but they are blurry for human eyes because they are learned by taking the average of various possible solutions for $I^G$.

This approach can be an efficient method under the assumption that the true distribution $p(I^G|I^J)$ of the lossless image $I^G$ corresponding to the lossy-compressed image $I^J$ is unimodal. 
However, $p(I^G|I^J)$ is ambiguous, because the mapping from the $I^J$ to the $I^R$ is one-to-many function which involves in quantization in each of the frequency channel.

Our goal is to create non-blurry $I^R$, which has sharp edges with vivid details. The proposed network tackles the problem in the frequency domain as the problem of estimating DCT coefficients $q$ of the $I^G$. 
Furthermore, instead of using the conventional MSE loss to directly estimate a single point $q$ in the space of DCT coefficients, a network is trained to estimate the distribution $p(q|I^J)$ of $q$ by minimizing the KL-divergence; 

\begin{equation} \label{ourporb}
\begin{aligned}
\theta &=\argmin_\theta D_{KL}(p(q|I^J)\|\hat{p}_{\theta}(q|I^J)). 
\end{aligned}
\end{equation}
Here, $\hat{p}_\theta$ is the estimated distribution by the network for the input $I^J$ with the parameter vector $\theta$.

The problem formulation of image restoration using KL-divergence has an advantage over that using MSE as follows:
Consider two images $I^G_1$ and $I^G_2$ result in the same JPEG image $I^J$. If a network is trained to directly estimate the target using two training samples $(I^J, I^G_1)$ and $(I^J, I^G_2)$, the network never learns because only a single point $I^R$ is outputted by the network. In this case, $I^R$ will be different from both targets $I^G_1$ and $I^G_2$. On the other hand, if a distribution is trained with the same samples, the network can learn the bimodal distribution which has two peaks at $I^G_1$ and $I^G_2$. If we take the maximal point of the distribution, one of the peaks will be selected and a perfect restoration is possible. 

In what follows, we treat $p(q|I^J)$ as the ground truth discrete distribution of the DCT coefficient classes in each frequency channel, and restore the image by using the information on the estimated distribution $\hat{p}(q|I^J)$ as the input to the traditional encoder-decoder neural network architecture.

\subsection{Overview of the proposed method}
Figure \ref{fig:ourmethod} shows the structure of the proposed method for image restoration. As can be seen in the figure, our compressed image restoration framework consists of three networks: a classifier, an encoder and a decoder. 
Through classification, classifier $C$ outputs a discrete distribution $\hat{p}$ which contains probability of per-patch frequency coefficient class for each frequency channel. Then the class with maximum probability is written in $\hat{y}$. 
From now on, 
$\hat{p}$ and $\hat{y}$ will be referred to as the \textit{class distribution map} and \textit{estimated class map}, respectively.
Mathematically, it becomes $\hat{p} = C(I^J)$ and $\hat{y} = \argmax_{k}{\hat{p}(k)}$. By a simple mapping $M$, the estimated class map $\hat{y}$ is further mapped to a map of real frequency coefficients $\hat{q}$,  \ie $\hat{q} = M(\hat{y})$.

The encoder $E$ takes $I^{J}$ to generate a feature map and
the decoder $D$ produces an output image $I^{R}$, which receives the detailed frequency information from the coefficients $\hat{q}$ estimated by the classifier as well as the output feature maps from the encoder:

\begin{equation} \label{overview}
\begin{aligned}
I^{R} &= D(E(I^{J}), \hat{q}). \\
\end{aligned}
\end{equation}

Instead of using the adversarial training scheme, the proposed method learns the network through typical supervised learning in the frequency domain. 
The cross entropy loss is used to train the classifier and the MSE loss is used to train the encoder and the decoder, thus the learning process is simple.

%Instead of using the $I^{G}$, we use the Laplacian image $I^{L}$ of the $I^G$ to highlight the details of an image. 
%In our method, the wider the range of the DCT coefficient, the larger becomes the quantization error. Using $I^{L}$ that has a smaller DCT coefficient range than $I^{G}$, we can reduce the quantization error and also can focus on detailed texture.

For practical reason, we define $q$ as DCT coefficient from Laplacian image $I^{L}$ of $I^{G}$, not from $I^{G}$.  this preprocessing can highlight the details of an image.  In our method, the wider the range of the DCT coefficient, the larger becomes the quantization error. Using $I^{L}$ that has a smaller DCT coefficient range than $I^{G}$, we can reduce the quantization error and also can focus on detailed texture.

%------------------------------------------------------------------------
\subsection{Classifier}

\begin{figure}
	\begin{center}
	\begin{subfigure}{0.43\textwidth} % width of left subfigure
		\includegraphics[width=\textwidth]{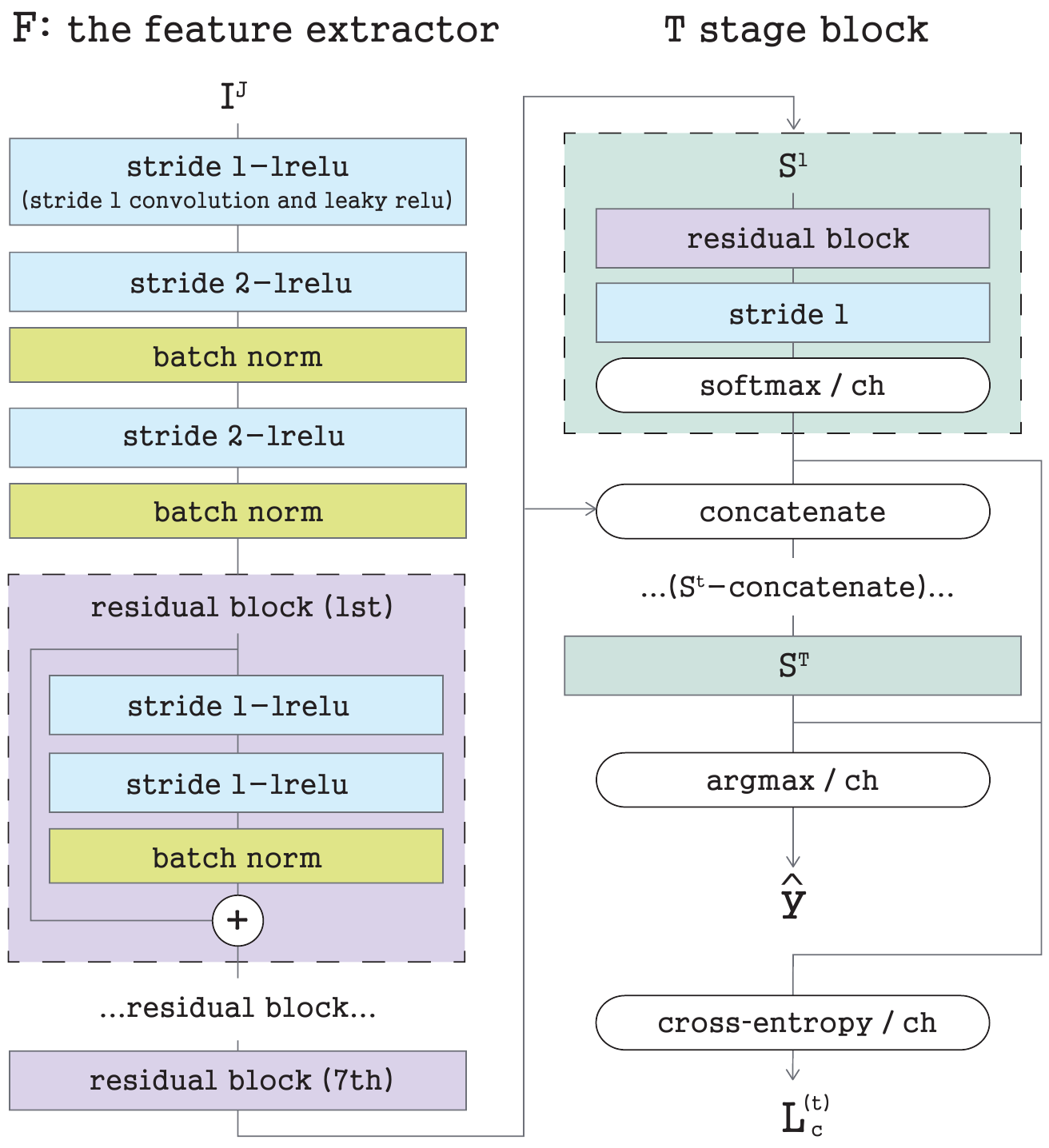}
		\caption{(a)} % subcaption
	\end{subfigure}
 	\vspace{1em} % here you can insert horizontal or vertical space
	\begin{subfigure}{0.43\textwidth} % width of right subfigure
		\includegraphics[width=\textwidth]{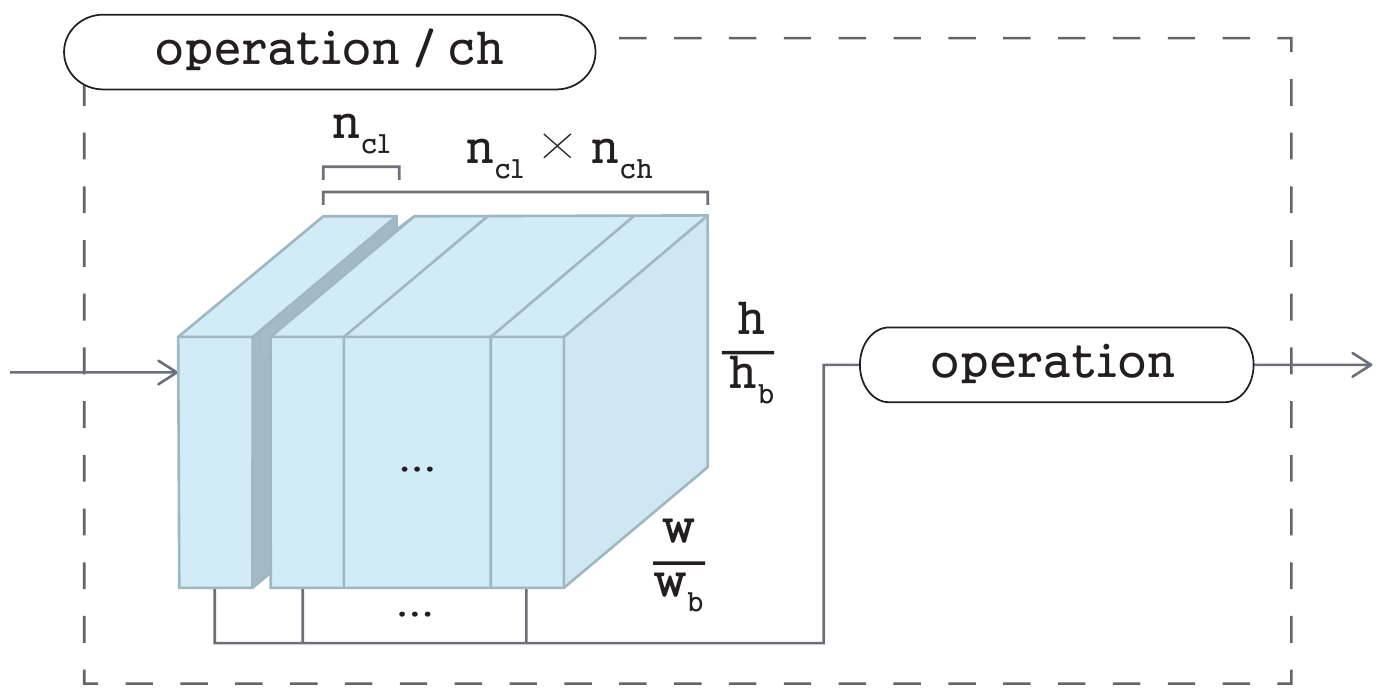}
		\caption{(b)} % subcaption
	\end{subfigure}
    \end{center}
	\caption{\textbf{The architecture of the classifier.} (a) is an illustration of the feature extractor $F$ and the multi stage block $S^{t}_{ch}$. In $F$ there are two stride 2 blocks because our $w_b$ and $h_b$ are set to 4.(b) represents per-channel operation of softmax, argmax, cross-entropy drawn as white box in (a). These operations are performed separately channel by channel.} 
    \label{fig:classifier}
\end{figure}

As shown in Figure \ref{fig:classifier}, the overall classifier network consists of a feature extractor $F$ and a multi-stage part composed of $T$ stage blocks.

The feature extractor part $F$ creates a feature map $f \in \mathbb{R}^{n_w \times n_h \times n_f}$ that contains spatial information of an input image $I^J \in \mathbb{R}^{w \times h \times c}$.
The parameters $n_w$ and $n_h$ are computed by $n_w = \frac{w}{w_b}$ and $n_h= \frac{h}{h_b}$, respectively, where $w_b$ and $h_b$ are the width and height of an image patch, $w$ and $h$ are the width and height of an input image, $c$ is the number of input channels which is typically 1 (gray) or 3 (RGB).

Each stage block $S^{t}, t = 1, \cdots, T,$ receives the feature map $f$ from the feature extractor. 
The first stage block $S^{1}$ generates a discrete class probability map $\hat{p}^{(1)}_{ch} \in \mathbb{R}^{n_w \times n_h \times\ n_{cl}}$ of the frequency coefficient corresponding to each frequency channel $ch$ of a local patch, where $n_{cl}$ is the number of frequency coefficient classes per channel
\footnote{If we use a $4 \times 4$ patch as a base block ($w_b = h_b = 4$), there are $16$ frequency channels. In our implementation, we set $n_{cl} = 7$, thus the output dimension of each stage is $112$ $(=16 \times 7)$ per an image patch.}.
Stage blocks $S^{t}, t = 2, \cdots,  T,$ concatenate the output $\hat{p}^{(t-1)} $of the previous block $S^{t-1}$ and the feature map $f$ from the feature extractor as an input:
\begin{equation} \label{stageeqn}
\begin{aligned}
(t = 1) \quad : \quad \hat{p}^{(1)}_{ch} &= S^{1}_{ch}(f) \\
(t > 1) \quad : \quad \hat{p}^{(t)}_{ch} &= S^{t}_{ch}(f,  \hat{p}^{(t-1)}). \\
\end{aligned}
\end{equation}
Here, $S^{t}_{ch}$ denotes the softmax output of the stage block $S^{t}$ for the frequency channel $ch \in \{1,\cdots, n_{ch} \}$ and the corresponding $\hat{p}^{(t)}_{ch}$ can be interpreted as a probabilistic estimation of the target class label $y_{ch}$. 

We design the multi-stage part to consider correlations between frequency channels and spatial blocks more deeply. 
Table \ref{table:mtstage} is the accuracy comparison of various levels of the stage blocks in the multi-stage classifier. As shown in the table, the stage 2 block shows a better result than the stage 1, but after the stage 2, the improvement is relatively low, so we decide to use only two stages $(T=2)$.

The classification result $\hat{y}_{ch} \in \mathbb{R}^{n_w \times n_h}$ is obtained by taking the index of the maximum element of the $\hat{p}^{(T)}_{ch}$ along the class axis as follows:
\begin{equation} \label{argmaxcls}
\begin{aligned}
\hat{y}_{ch} = \argmax_{k}(\hat{p}^{(T)}_{ch}(k)), \quad k \in \{1, \cdots, n_{cl} \}. \\
\end{aligned}
\end{equation}
The class $\hat{y}_{ch}$ estimated by the classifier is converted back to the final DCT coefficient value $\hat{q}_{ch} = M_{ch}(\hat{y}_{ch})$ to represent the real information. Here, $M_{ch}(\hat{y}_{ch})$ is our class-to-coefficient mapping.

We use the cross entropy loss to train the classifier. The classification loss at each stage $L^{(t)}_{c}, t=1, \cdots, T,$ is first calculated as the average of the cross entropy losses in all the spatial blocks and frequency channels using the class distribution map $\hat{p}^t$ and the ground truth class map $y$.
Then, these are averaged to define the final classification loss:
\begin{equation} \label{lclass}
\begin{aligned} 
L_{c} &= \frac{1}{T}\sum_{t=1}^{T}L^{(t)}_{c}.
\end{aligned}
\end{equation}

The cross entropy loss is closely related to the KL-divergence and plays a very important role in our method in that it matches the predicted distribution $\hat{p}$ with the ground truth frequency coefficient class $y$.

\begin{table}[t]
\begin{center}
\label{table:multistage}
\begin{tabular}{| l | l | l | l |}
\hline
  		 & 	stage 1	&   stage 2	&  stage 3  \\ 
\hline
accuracy &  0.3094  &  0.3169   &  0.3178  \\ 
\hline
\end{tabular}
\caption{Classification accuracy of each stage evaluated on a three-stage classifier, trained on 96 $\times$ 96 luminance images.
}
\label{table:mtstage}
\end{center}
\end{table}

%------------------------------------------------------------------------
\subsection{Encoder and Decoder}
Like the classifier, our encoder-decoder architecture consists of several residual blocks and convolution layers.
The structure of the encoder is exactly same as the feature extractor($F$) of the classifier, and that of the decoder is exactly symmetric with the encoder except that there is no activation function after 
the output convolution layer.
In the decoder, upsampling (convolution - pixel shuffle \cite{shi2016pixelshuffle} - leaky ReLU \cite{maas2013rectifier}), instead of downsampling, is conducted.
The encoder $E$ in Fig. \ref{fig:ourmethod} produces a feature map of an input image, while the decoder $D$ takes the output of the encoder $E$ and the result of the classifier $C$ together as an input to produce an output image.

The output of the classifier $\hat{q}$ has frequency information that can concatenate to the feature maps induced by the input image $I^J$. 
In order for $\hat{q}$ to be used appropriately to produce the output image, the decoder needs to learn the mapping from the frequency domain to the pixel domain. 
If we use estimated $\hat{q}$ in the training of $D$ with the general pixel-wise distance loss, 
the decoder has a tendency to ignore $\hat{q}$ and mostly rely on the feature map induced by the input image $I^J$. This is due from the fact that $\hat{q}$ does not have a perfect information on the target image $I^G$ because of the classification error in $C$ and this small but incorrect error in the frequency domain can cause large errors in each pixel value for the entire patch.

\begin{figure}
 	\centering
	\begin{center}
      \begin{subfigure}{0.20\textwidth} % width of left subfigure
          \includegraphics[width=\textwidth]{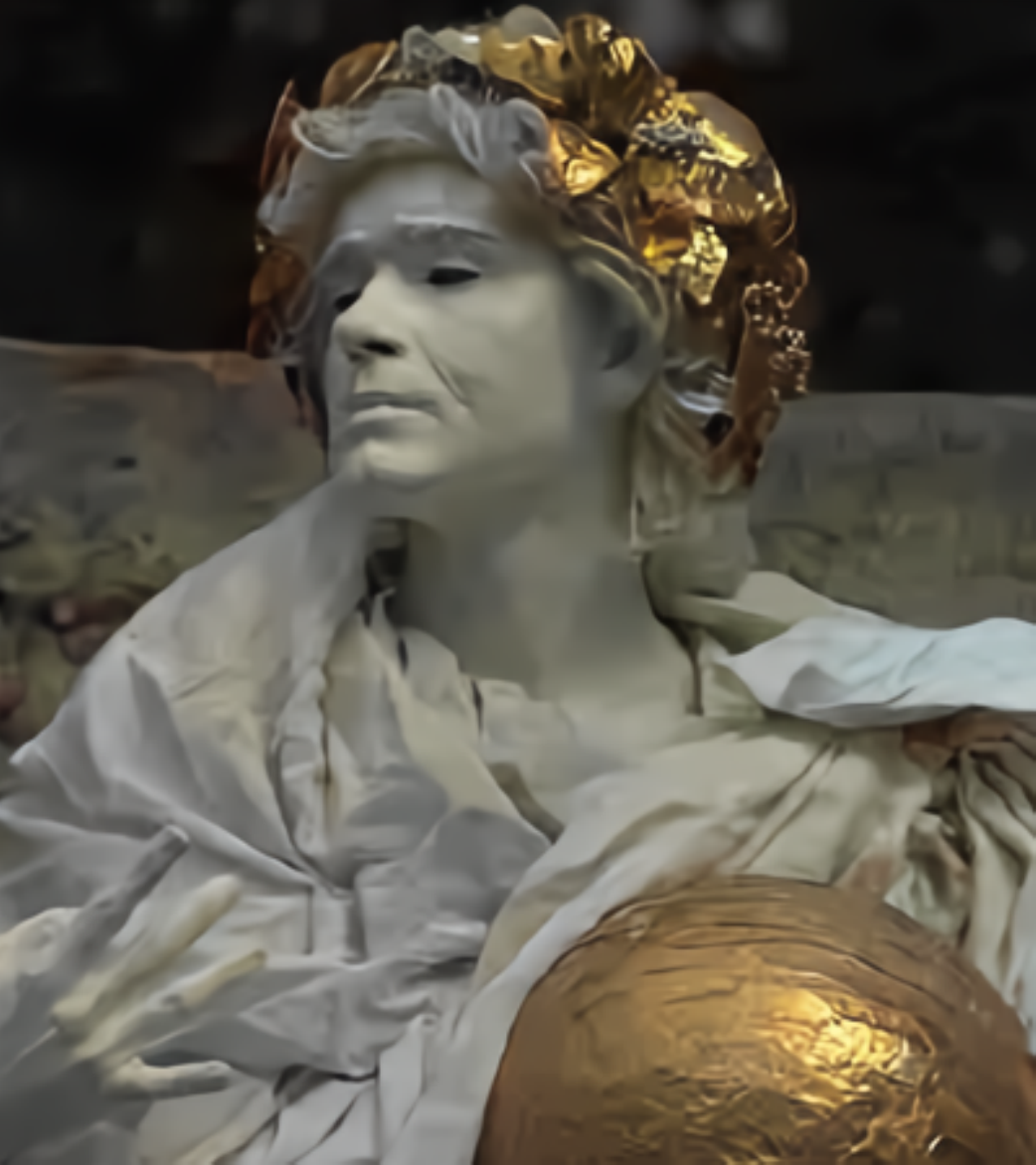}
      \end{subfigure}
      \qquad
      \begin{subfigure}{0.20\textwidth} % width of right subfigure
          \includegraphics[width=\textwidth]{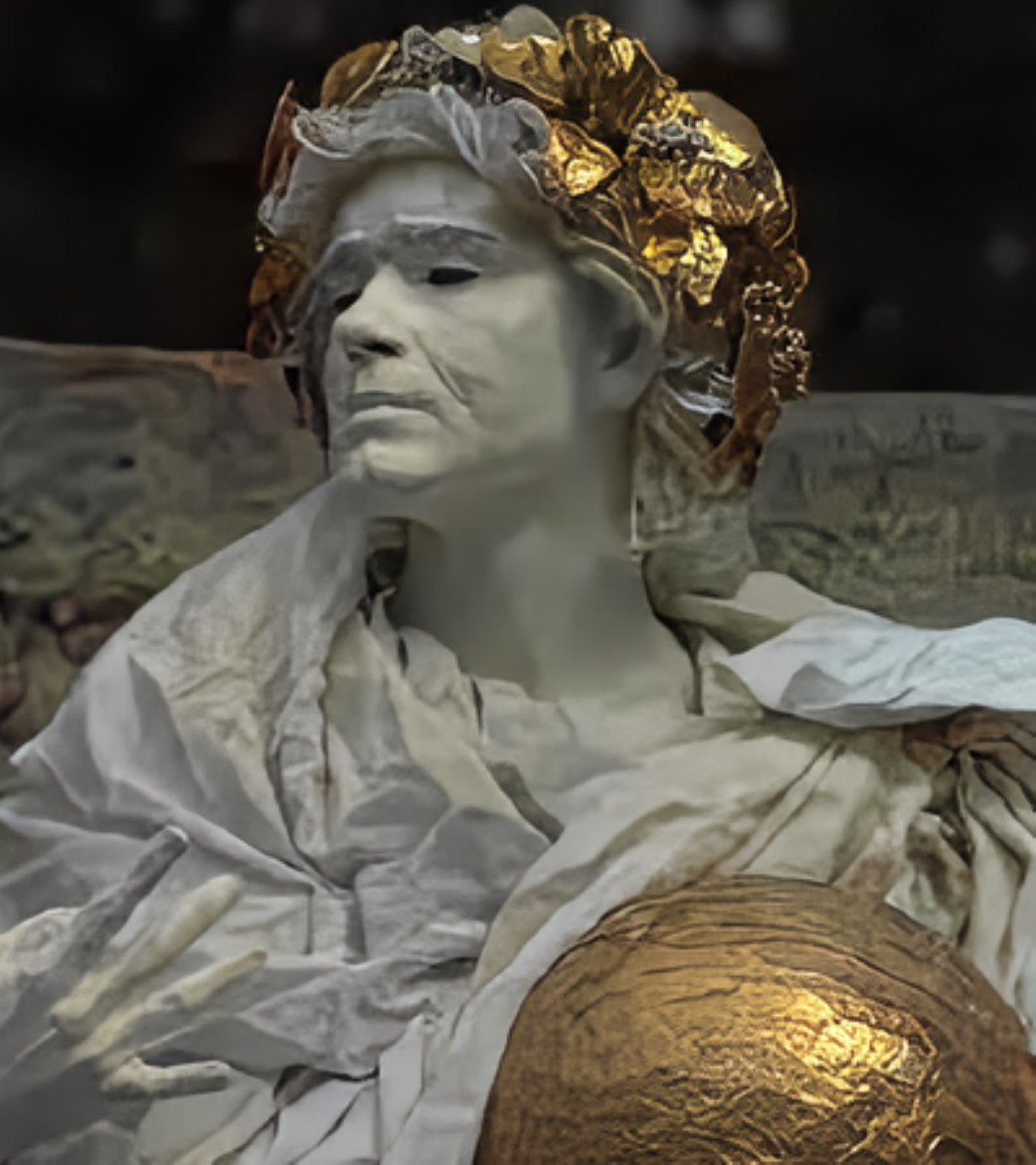}
      \end{subfigure}
    \end{center}
        \vspace{-2mm}
	\caption{The effect of using ground truth $q$ instead of classifier output $\hat{q}$ for training the encoder-decoder. Both images are using the same classifier, but the encoder-decoder is trained by different loss. The image on the right is the output of the network trained by equation (\ref{eqn:lreconst}) using $q$, while the left one is the result of the network trained by $\hat{q}$ instead of $q$ in equation (\ref{eqn:lreconst}). In test time, both images are generated using $\hat{q}$. The right one has more vivid detail and lively edges.}
    \label{fig:comparison-est-gt}
\end{figure}

However, if we use the exact frequency information of the target image, the decoder can learn the mapping from the frequency to the pixel domain correctly.
Therefore, in the training phase, instead of using the imperfect frequency information $\hat{q}$, we use $q$, the actual frequency information obtained from the target image $I^G$, in the reconstruction loss, as follows: 
\begin{equation}\label{eqn:lreconst}
L_{r}  =   
\frac{1}{wh}\sum_{x=1}^{w}\sum_{y=1}^{h}\|
\mathit{D}(\mathit{E}(I^{J}), \mathit{q})_{x,y} - I^{G}_{x,y}\|^{2},
\end{equation}
where the subscript $x,y$ indicates the pixel location. The reconstruction loss $L_{r}$ is back-propagated to the encoder and decoder and does not affect the classifier.

%Figure \ref{fig:comparison-est-gt} compares the effect of using ground truth $q$ instead of the classifier output $\hat{q}$.
As shown in the Figure \ref{fig:comparison-est-gt}, the image reconstructed by the network trained with ground truth $q$ has more vivid and natural detail. 
We have calculated the mean absolute values of convolution filters in the first layer of the decoder for both cases of using $q$ and $\hat{q}$ in the training. Then, the elements corresponding to the feature vector $f$ from the encoder and the ones corresponding to the DCT coefficients $q$ or $\hat{q}$ from the classifier are summed to yield $w_f$ and $w_q$, respectively. In both networks, $w_f$ is bigger than $w_q$, but the difference of this value $w_f-w_q$ of using $\hat{q}$ is about 1.74 time bigger than that using $q$. Therefore, we can assume that the decoder trained using $\hat{q}$ ignores the features from the classifier much.

Since $q$ contains DCT coefficients of the target image $I^{G}$, it is used only for training. In the test phase, the output image $I^{R}$ is generated by using the estimated $\hat{q}$. 

%------------------------------------------------------------------------

\begin{figure*}[th]	
% visual analysis figures
	\centering
%%%%%% COL 1
    \begin{subfigure}[t]{0.18\textwidth}
        \begin{subfigure}{\textwidth} % width of left subfigure
            \includegraphics[width=\textwidth]{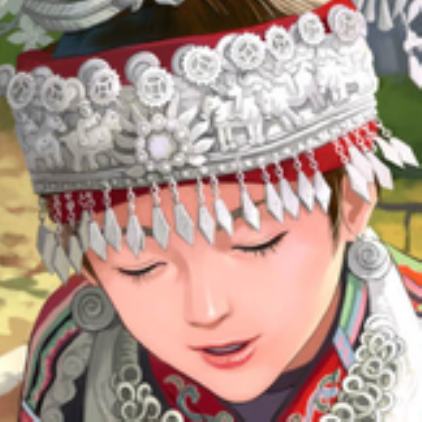}
        \end{subfigure}
    
 		\vspace{1mm} % here you can insert horizontal or vertical space
%         \vspace*{0.5mm} % here you can insert horizontal or vertical space
        \begin{subfigure}{\textwidth} % width of right subfigure
            \includegraphics[width=\textwidth]{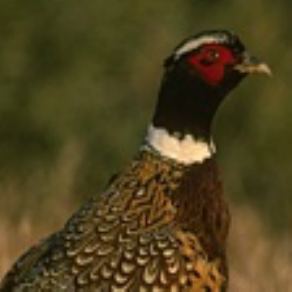}
        \end{subfigure}
		\caption{Original}
        
        \vspace{2mm} % here you can insert horizontal or vertical space
        \begin{subfigure}{\textwidth} % width of right subfigure
            \includegraphics[width=\textwidth]{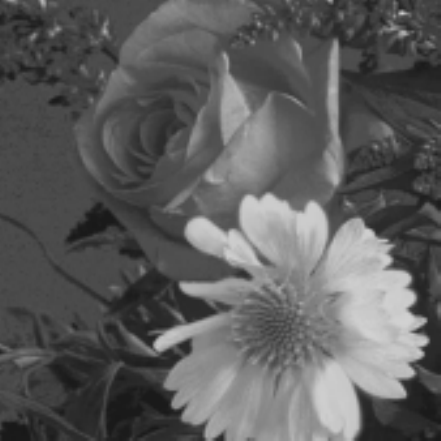}
        \end{subfigure}
		\caption{Original}
    \end{subfigure}%
    ~
%%%%%% COL 2
    \begin{subfigure}[t]{0.18\textwidth}
    	\begin{subfigure}{\textwidth} % width of left subfigure
            \includegraphics[width=\textwidth]{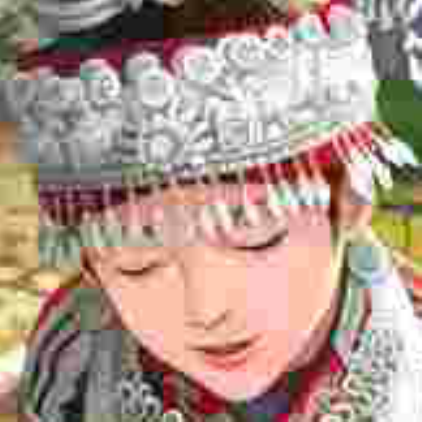}
%   			\vspace*{0.5mm} % here you can insert horizontal or vertical space
        \end{subfigure}
    
 		\vspace{1mm} % here you can insert horizontal or vertical space
        \begin{subfigure}{\textwidth} % width of right subfigure
            \includegraphics[width=\textwidth]{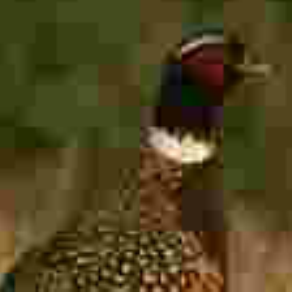}
        \end{subfigure}
    	\caption{JPEG (QF 10)}
        
        \vspace{2mm} % here you can insert horizontal or vertical space
        \begin{subfigure}{\textwidth} % width of right subfigure
            \includegraphics[width=\textwidth]{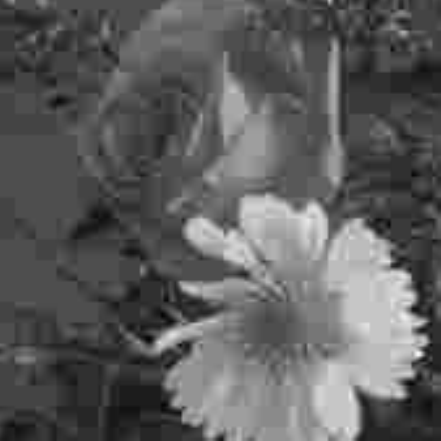}
        \end{subfigure}
		\caption{JPEG (QF 10)}
    \end{subfigure}
	~
%%%%%% COL 3
    \begin{subfigure}[t]{0.18\textwidth}
    	\begin{subfigure}{\textwidth} % width of left subfigure
            \includegraphics[width=\textwidth]{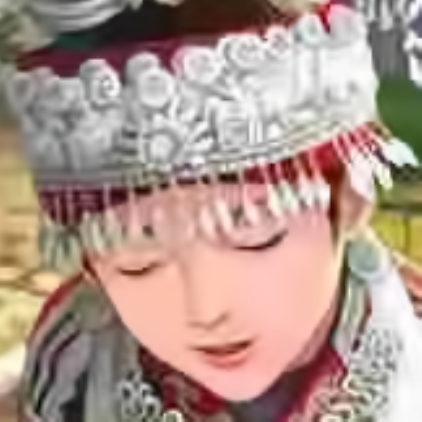}
%   			\vspace*{1mm} % here you can insert horizontal or vertical space
        \end{subfigure}
    
 		\vspace{1mm} % here you can insert horizontal or vertical space
        \begin{subfigure}{\textwidth} % width of right subfigure
            \includegraphics[width=\textwidth]{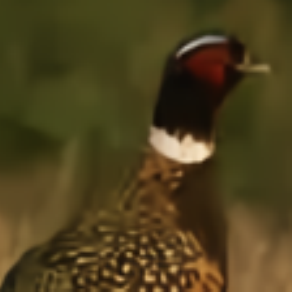}
        \end{subfigure}
        \caption{SA-DCT\cite{foi2007pointwise}}
        
        \vspace{2mm} % here you can insert horizontal or vertical space
        \begin{subfigure}{\textwidth} % width of right subfigure
            \includegraphics[width=\textwidth]{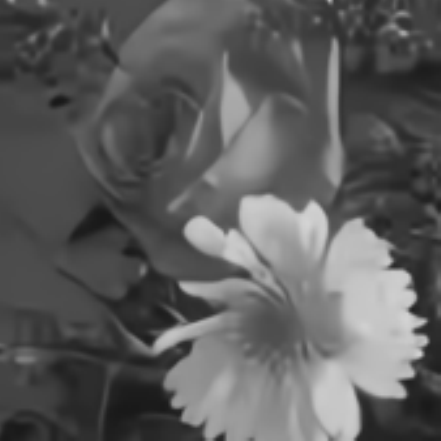}
        \end{subfigure}
		\caption{AR-CNN\cite{arcnn}}
    \end{subfigure}
    ~
%%%%%% COL 4
    \begin{subfigure}[t]{0.18\textwidth}
    	\begin{subfigure}{\textwidth} % width of left subfigure
            \includegraphics[width=\textwidth]{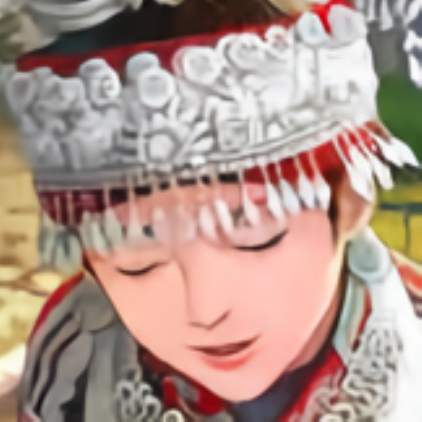}
%   			\vspace*{1mm} % here you can insert horizontal or vertical space
        \end{subfigure}
    
 		\vspace{1mm} % here you can insert horizontal or vertical space
        \begin{subfigure}{\textwidth} % width of right subfigure
            \includegraphics[width=\textwidth]{figure/bird-ed.pdf}
        \end{subfigure}
        \caption{ED}
        
        \vspace{2mm} % here you can insert horizontal or vertical space
        \begin{subfigure}{\textwidth} % width of right subfigure
            \includegraphics[width=\textwidth]{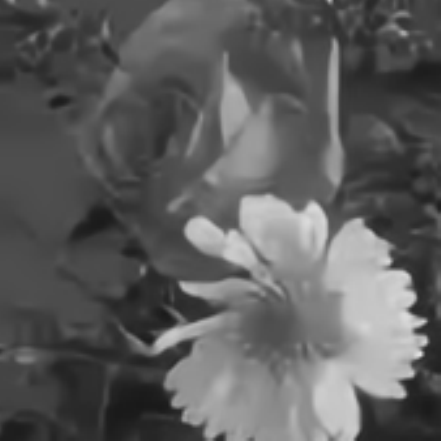}
        \end{subfigure}
		\caption{ED}
    \end{subfigure}
    ~
%%%%%% COL 5
    \begin{subfigure}[t]{0.18\textwidth}
    	\begin{subfigure}{\textwidth} % width of left subfigure
            \includegraphics[width=\textwidth]{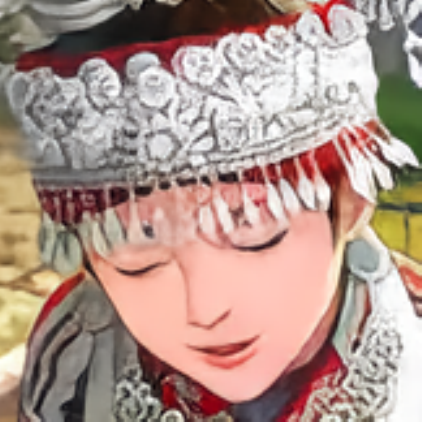}
%   			\vspace*{1mm} % here you can insert horizontal or vertical space
        \end{subfigure}
    
 		\vspace{1mm} % here you can insert horizontal or vertical space
        \begin{subfigure}{\textwidth} % width of right subfigure
            \includegraphics[width=\textwidth]{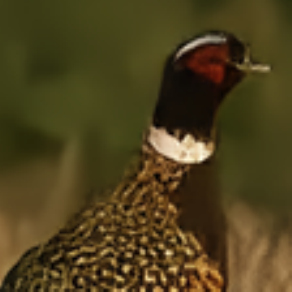}
        \end{subfigure}
        \caption{Ours (CED-GT)}
        
        \vspace{2mm} % here you can insert horizontal or vertical space
        \begin{subfigure}{\textwidth} % width of right subfigure
            \includegraphics[width=\textwidth]{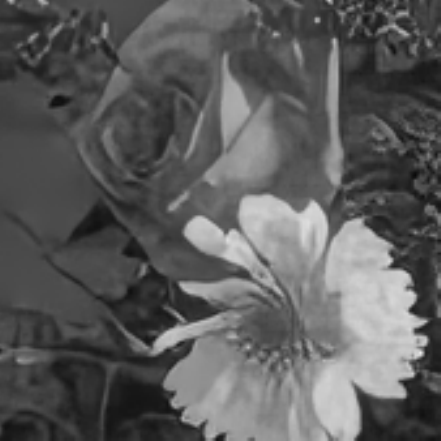}
        \end{subfigure}
		\caption{Ours (CED-GT)}
    \end{subfigure}
    \caption{Comparison of restored result according to loss function used in learning. The first and second rows are image extracted on RGB channel. The third low is result on Y channel. The QF in jpeg images means the quality factor. In every samples, while the others' results look blurred, ours are most clear with high frequency details and edges being the closest to the ground truths. }
    \label{fig:comparison}
\end{figure*}

\section{Experiments}
\label{sec:experiment}
\subsection{Experiment setting}

For all the experiments in this section, our baseline (ED) uses the encoder($E$) and the decoder($D$) networks without the classifier, trained using only the corresponding reconstruction loss.
The proposed method, CED-GT, is trained and tested with the classifier ($C$) and the reconstruction loss. In this case, the ground truth DCT coefficient map $q$ is used for the training of $E$ and $D$. 
For comparison, we also trained $E$ and $D$ using the estimated DCT coefficient map $\hat{q}$ and name this model as CED-EST. 
For the classifier, the number of stage modules ($T$) is set to two.

For our training data, the test set of ILSVRC 2015 \cite{ILSVRC15} which contains $10^5$ images was chosen. For data augmentation, these images were horizontally flipped and total $2 \times 10^5$ images were used for training. For test data, and the validation sets of BSDS500 \cite{bsds500} and LIVE1 \cite{sheikh2016live} were used.
For fair comparison with previous works, we used MATLAB JPEG encoder as others did. Depending on the experiment, the quality factor of 10 or 20 was used.

At the training phase, the training images were resized to $96 \times 96$ or $128 \times 128$ depending on the dimension of the network. In the test phase, full-sized images were directly restored using the fully convolutional property of the network.

To generate a label for the training of the classifier network, we first extracted the luminance channel Y from the YCbCr format of the target image and then took the Laplacian to highlight the image detail.
Then, the image was divided into $4 \times 4$ patches. By performing DCT on each patch, 16 channels of DCT coefficients were obtained, and then each of the 16 coefficients was labeled to one of the 7 classes. To prohibit class imbalance problem, the DCT coefficient spaces are evenly separated such that each class bin has the same number of training samples.

\begin{figure*}[t]
    \centering
    \begin{subfigure}[t]{0.23\textwidth}
        \centering
        \includegraphics[width=\linewidth]{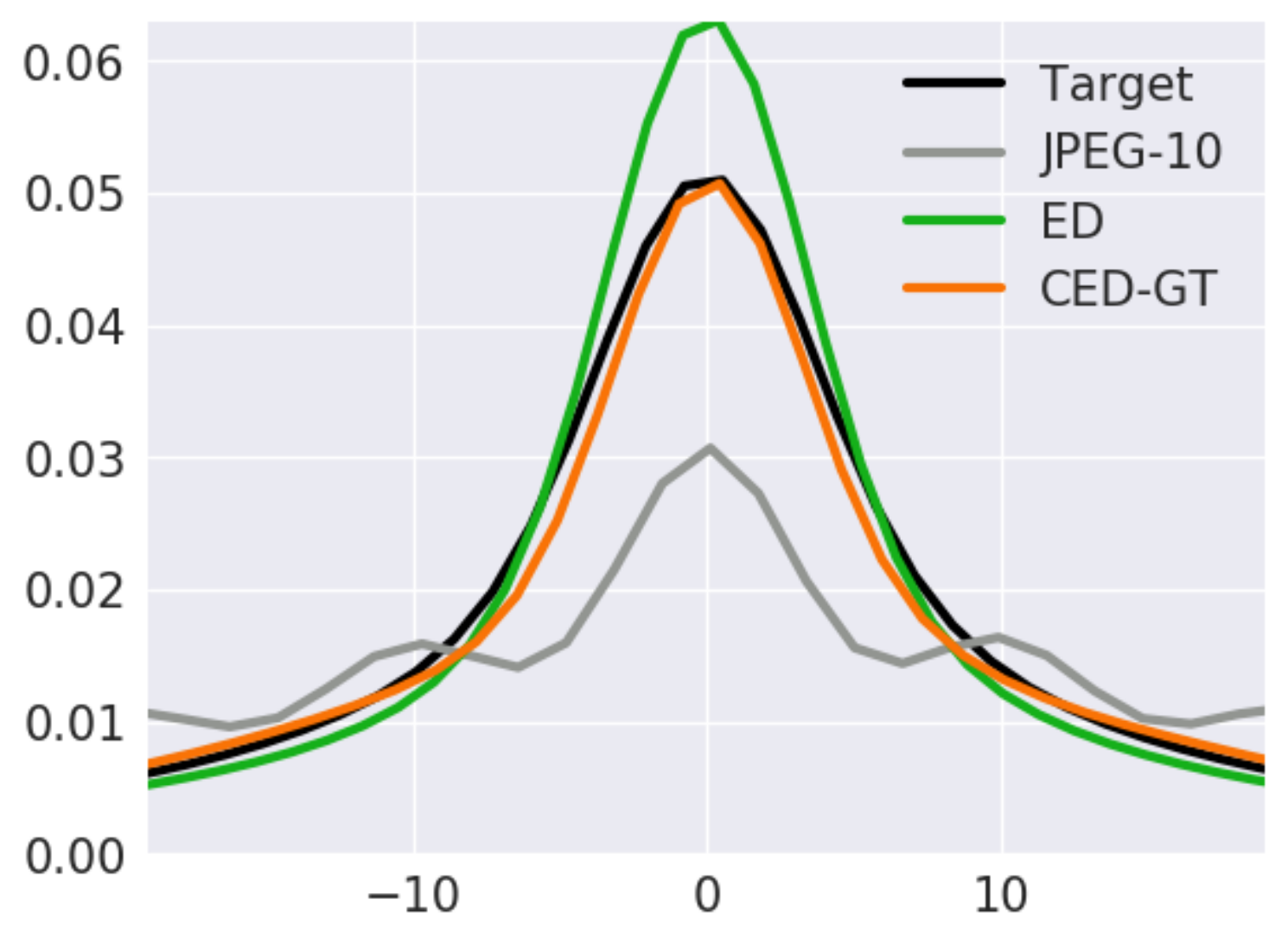}
        \caption{\specialcell{1-1\\(lowest frequency)}}
    \end{subfigure}%
    ~
    \begin{subfigure}[t]{0.23\textwidth}
        \centering
        \includegraphics[width=\linewidth]{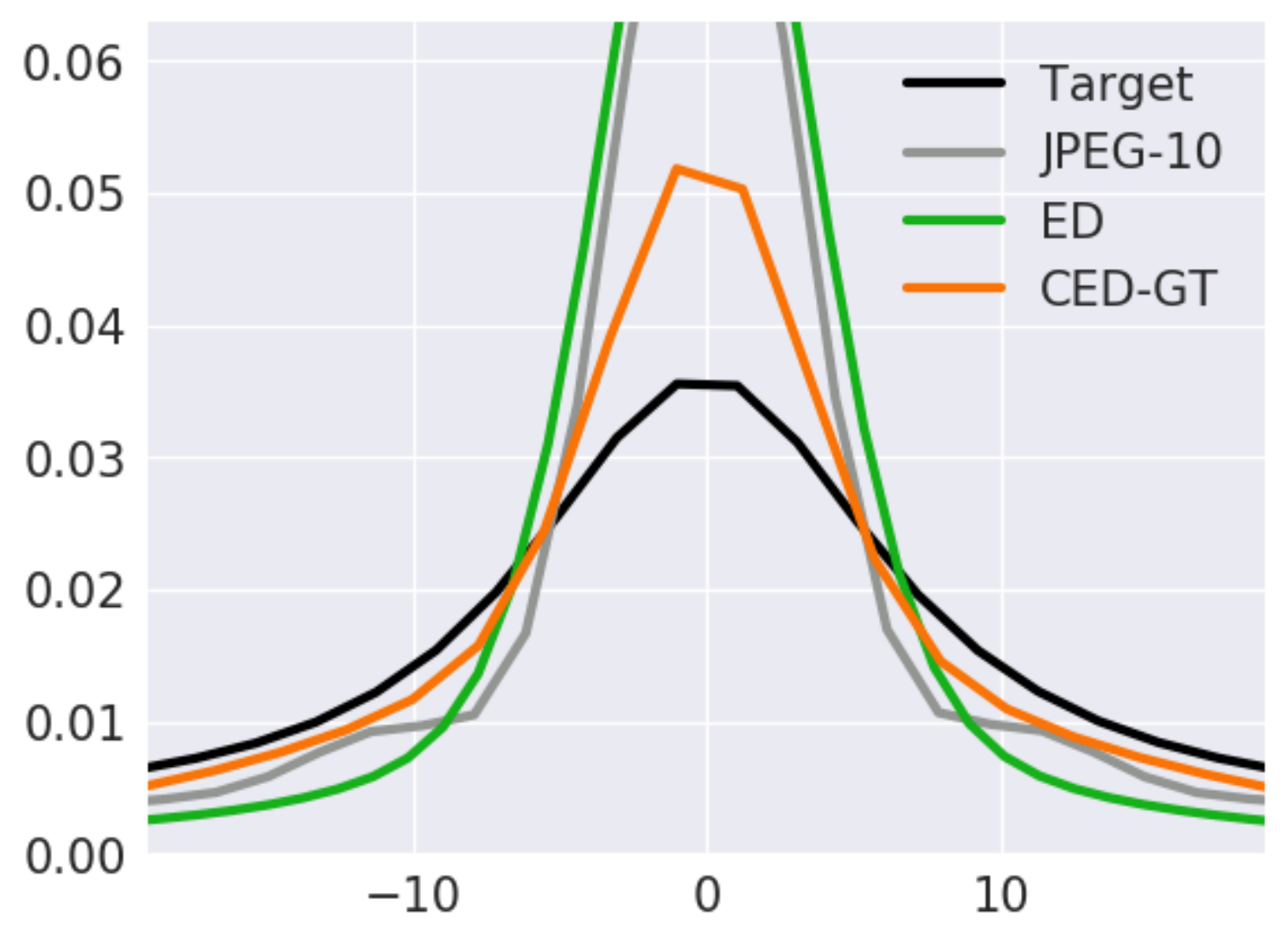}
        \caption{4-4}
    \end{subfigure}
    ~
    \begin{subfigure}[t]{0.23\textwidth}
        \centering
        \includegraphics[width=\linewidth]{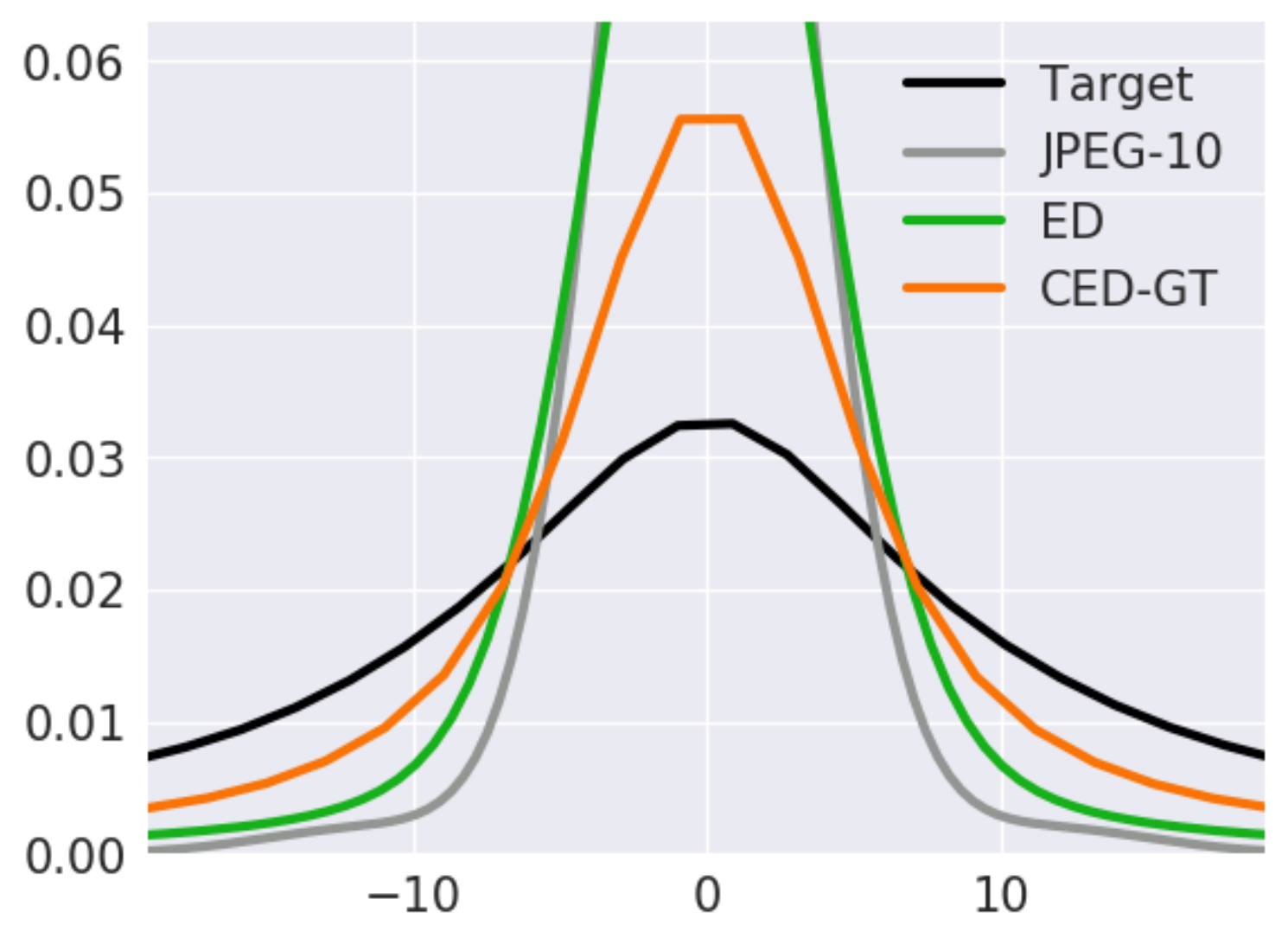}
        \caption{6-6}
    \end{subfigure}
    ~
    \begin{subfigure}[t]{0.23\textwidth}
        \centering
        \includegraphics[width=\linewidth]{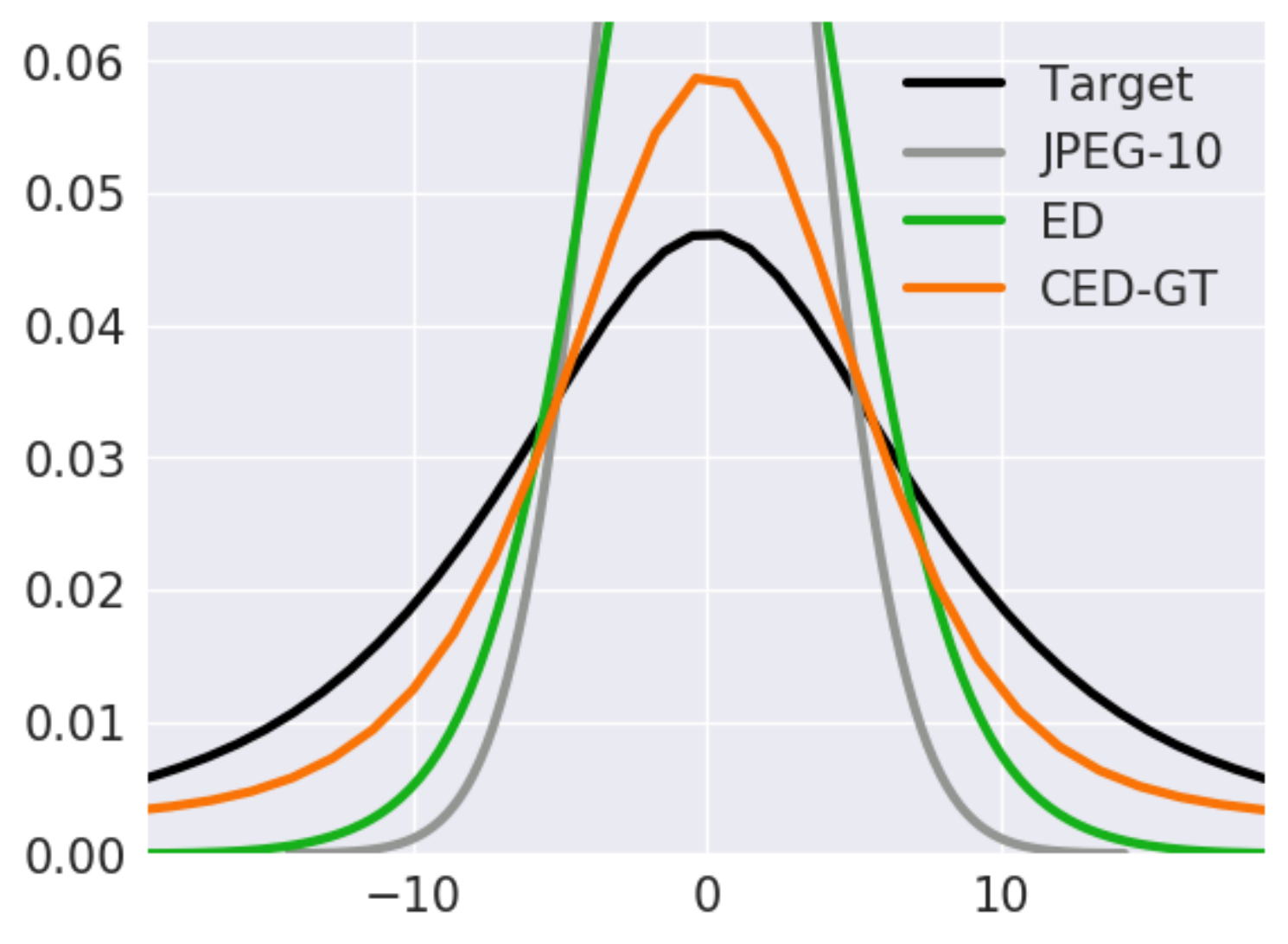}
        \caption{8-8\\(highest frequency)}
    \end{subfigure}
    \caption{The distribution of some DCT coefficients of laplacian images, obtained through 2D-DCT of 8x8 image patches in LIVE1 dataset. The 1-1 means first row, first column in frequency table. we compare the ground truth, JPEG (quality factor 10) and restored image (using baseline ED and our methods CED-GT). The distributions of our method are more similar to the ground truths, yet baseline method ED is similar to the JPEG. 
    }
    \label{fig:ex_freq_distribu}
\end{figure*}

\subsection{Qualitative result}
Figure \ref{fig:comparison} is the results of the proposed method.
The proposed method, CED-GT, is compared with the baseline ED and SA-DCT \cite{foi2007pointwise}.
The grayscale model of AR-CNN \cite{arcnn} by the author, which is publicly available is also used for comparison.
Compared to the source image JPEG-10, the output of ED 
is closer to the ground truth image, but the details still look burred.
The proposed CED-GT shows the best result, hardly distinguishable from the ground truth with bare eyes.
Our method removes a block boundary effects, as well as a ringing artifacts. For the high frequency detailed surfaces, such as feather, flowers, the results of ours have more visually plausible texture compared with those of others. Especially, the blurry edges that the ED method generates turn to very vivid ones in ours.
Our method does not make such a texture randomly. For the low frequency region such as flat surface, our method does not make such a rough texture.
From this, we can conjecture that the network knows where to generates the detail or not. Furthermore, the network does not just sprinkle the pattern, rather it generates visually natural edges.

\subsection{Quantitative result}
\textbf{Frequency distribution}
To show our method successfully restores the frequency detail, the distribution of DCT coefficients are compared between methods.
To focus on detail, we extracted DCT coefficients on Laplacian images.
Figure \ref{fig:ex_freq_distribu} shows the distribution of frequency coefficients for various frequency channels. The distribution was obtained by applying 2D-DCT on $8 \times 8$ image patches of LIVE1 dataset \cite{sheikh2016live}.
Except the DC component (1-1 channel), the DCT coefficients of target images are widely distributed, whereas the distribution of JPEG image is very narrow and concentrated near zero. This tendency is strengthened with increasing frequency. This is because the JPEG compression absolutely removes its high frequency components in most of the time. 
As shown in Figure \ref{fig:ex_freq_distribu}, the distribution of the images that were restored by the proposed CED-GT is more similar to that of the target image, compared to the ED baseline. On the other hand, the distribution of the ED is similar to that of JPEG images, especially at high frequencies.

\textbf{BEF and quantitative metric} For quantitative comparison, we adopt the well-known traditional metrics PSNR, PSNR-B~\cite{yim2011quality}, and SSIM~\cite{ssim} scores. However, according to the previous state of the art works using GAN \cite{srgan, onetomany, l.galteri2017gan}, this measurement may not be perfectly related to human vision. It is clear that the images generated by their methods are visually better for human than the plain MSE-based images, but their PSNR scores are generally lower than those of the MSE-based methods.
The image restoration problem can be seen as estimating a solution that is plausible to human eyes, not perfectly restoring an unknown target pixel-by-pixel. Therefore, if the human visual evaluation is satisfied, it is acceptable to have a lower value for MSE-related measure. \\

\begin{table*}[t]
\centering
%\begin{small}
\resizebox{0.65\linewidth}{!}{
\begin{tabular}{|c|c|c|c|c|c|c|c|c|c|}
\hline
\multirow{2}{*}{QF}	&\multirow{2}{*}{Method}		& \multicolumn{4}{|c|}{LIVE1}
													& \multicolumn{4}{|c|}{BSDS500}\\ \cline{3-10}
					&					& \specialcell{BEF\\/MSE}$>$ 	& PSNR 		& PSNR-B	& SSIM 	
										& \specialcell{BEF\\/MSE}$>$ 	& PSNR 		& PSNR-B	& SSIM\\
\hhline{==========}
\multirow{6}{*}{10}	
& JPEG							      & 0.754 		& 27.77 		& 25.33 		& 0.791
									  & 0.824 		& 27.58 		& 24.97 	    & 0.769 \\
& *AR-CNN\cite{arcnn}				  & 0.094 		& 29.13 		& 28.74 		& 0.823
									  & 0.086 		& 28.74 		& 28.38 		& 0.796 \\
& Galteri-MSE\cite{l.galteri2017gan}  & 0.067 		& 29.41 		& 29.13			& 0.832
									  & 0.089 		& 28.93			& 28.56 		& 0.805 \\
& *Galteri-MSE\cite{l.galteri2017gan} & 0.084	 	& 29.45			& 29.10 		& 0.834
									  & 0.102 		& 29.03 		& 28.61 		& 0.807 \\
& *Galteri-GAN\cite{l.galteri2017gan} & 0.148  		& 27.29 		& 26.69 		& 0.773
									  & 0.178		& 27.01 		& 26.30			& 0.746 \\
& ED	 							  & 0.074 		& 29.40			& 29.09 		& 0.833
									  & 0.094 		& 28.96 		& 28.57 		& 0.806 \\
& CED-EST 						      & 0.076 		& 29.40 		& 29.08 		& 0.832
									  & 0.094 		& 28.95 		& 28.56 		& 0.805 \\
& CED-GT	 					      & \textbf{0.007}& 26.54 		& 26.51 		& 0.767
									  & \textbf{0.007}& 26.00 		& 25.97 		& 0.731 \\ \hline
\multirow{6}{*}{20}
& JPEG								  & 0.778		& 30.07 		& 27.57 		& 0.868 
									  & 0.884 		& 29.72 		& 26.97 		& 0.852 \\
& *AR-CNN\cite{arcnn}				  & 0.178 		& 31.40 		& 30.69 		& 0.890
									  & 0.180 		& 30.80 		& 30.08 		& 0.868 \\
& Galteri-MSE\cite{l.galteri2017gan}  & 0.122		& 31.70 		& 31.20 		& 0.896
									  & 0.180		& 31.09 		& 30.37 		& 0.876 \\
& *Galteri-MSE\cite{l.galteri2017gan} & 0.125 		& 31.77 		& 31.26 		& 0.896
									  & 0.180		& 31.20 		& 30.48 		& 0.832 \\
& *Galteri-GAN\cite{l.galteri2017gan} & 0.059 		& 28.35 		& 28.10 		& 0.817 
									  & 0.740 		& 28.07 		& 27.76 		& 0.794 \\
& ED	 							  & 0.132		& 31.68 		& 31.14 		& 0.895
									  & 0.189 		& 31.08 		& 30.33 		& 0.875 \\
& CED-EST 						      & 0.127		& 31.65 		& 31.13 		& 0.895
									  & 0.180		& 31.04 		& 30.32 		& 0.875 \\
& CED-GT	 					      & \textbf{0.002}& 29.33 		& 29.32 		& 0.854
									  & \textbf{0.009}& 28.62 		& 28.58 		& 0.825 \\ \hline
\end{tabular}}
%\end{small}
\caption{The lower bound of mean BEF/MSE and other quantitative result on LIVE1 and BSDS500. All the experiments are done with luminance images. The BEF/MSE of our CED-GT methods are quite lower than others, which means block artifact effect is very low. For the Galteri \etal \cite{l.galteri2017gan}, the asteric (*) marked results are brought directly from their paper, and the other ones are reproduced by ourselves. The results of AR-CNN \cite{arcnn} are also brought from \cite{l.galteri2017gan}}
\label{table:ex_quantity_result}
\end{table*}

For the above reason, the PSNR score does not fit for the assessment of modern state-of-the-art image restoration methods. Inspired by \cite{yim2011quality}, we also evaluate the generated images in the BEF metric, which is defined by the difference in MSEs between block boundary pixels ($D_B(I^R)$) and non-block boundary pixels ($D^C_B(I^R))$):factor
\begin{equation} \label{perc_quant}
\begin{aligned}
\mathbf{BEF}(I^R) &= \eta (D_B(I^R) - D^C_B(I^R)), 
\end{aligned}
\end{equation}
where $\eta$ is a constant that depends only on the image and block sizes. If the boundary artifact effect is zero, there is clearly no reason that $D_B(I^R)$ is different with $D^C_B(I^R)$. So the lower BEF indicates lower block boundary effect. 
If the BEF score is normalized by MSE of an entire image (\textbf{BEF/MSE}), it can be a good measure of how much boundary effect is removed. However, as we do not know the each image's BEF of other works, we can only derive the lower bound of mean BEF/MSE via Jensen's inequality, using mean PSNR and mean PSNR-B that are provided.

Table \ref{table:ex_quantity_result} shows a quantitative comparison of various algorithms on LIVE1 and BSDS500 datasets. All the quantitative experiments were done using the luminance (Y) channel of the original YCbCr images. The quality factors of JPEG compression were 10 and 20. The proposed methods were compared with the methods presented by Galteri \etal \cite{l.galteri2017gan}. The MSE version in \cite{l.galteri2017gan} was implemented and trained by ourselves but we could not reproduce the GAN version. In the table, the results with $*$ mark are directly from the paper \cite{l.galteri2017gan}. Considering the MSE score we reproduce (Galteri-MSE) is not that much different from their report (*Galteri-MSE), we believe this comparison is plausible.

As shown in Table \ref{table:ex_quantity_result} the PSNR score of the proposed CED-GT are lower than that of the other methods, but still its output images are well restored and visually better than other methods. The BEF/MSE score of CED-GT is much lower ($\times5\sim\times10$) than that of the other methods. Note that the GAN-based method also has high BEF values which means that block boundary effect is not treated well.

\subsection{High level tasks: detection and segmentation}
If the image is successfully restored, so that the image is abundant in high frequency details, we can assume that the image can yield better performance on high level vision tasks, such as object detection and semantic segmentation. We tested Faster RCNN \cite{renNIPS15fasterrcnn} and FCN-8s \cite{shelhamer2017fully} as a benchmark algorithms for object detection and semantic segmentation, respectively, and compared their scores on images generated by various methods. Note that ARCNN is tested only on grayscale images. 

Table \ref{table:det_seg} shows the result of detection and segmentation on VOC 2007 \cite{voc}. We subsampled the VOC 2007 to collect sample that labeled to both detection and segmentation task. All the experiments on Table \ref{table:det_seg} is performed on this subset.For both detection and segmentation tasks, the proposed CED-GT obtained the highest scores of mAP and mIoU.
\begin{table}[h]
\centering
\begin{small}
\begin{tabular}{|c|c|c|c|c|}
\hline
\multirow{3}{*}{Method}				& \multicolumn{2}{|c|}{Detection}			& \multicolumn{2}{|c|}{Segmentation}	\\ \cline{2-5}
									&\specialcell{mAP\\(RGB)} &\specialcell{mAP\\(Y)}&\specialcell{mIoU\\(RGB)}&\specialcell{mIoU\\(Y)}			\\ \hhline{=====}
JPEG (QF 10)								& 0.359					& 0.292				& 0.414				& 0.311 			\\
SA-DCT\cite{foi2007pointwise}		& 0.485					& 0.426				& 0.456				& 0.363				\\
ARCNN\cite{arcnn}					& -						& 0.429				& -					& 0.375				\\
Galteri-MSE\cite{l.galteri2017gan}	& 0.519					& 0.438				& 0.462				& 0.378				\\
ED									& 0.525					& 0.437				& 0.471				& 0.380				\\
CED-EST								& 0.526					& \textbf{0.440}	& 0.474				& 0.384				\\
CED-GT								& \textbf{0.550}		& \textbf{0.440}	& \textbf{0.475}	& \textbf{0.389}	\\ \hline
Original							& 0.705					& 0.637				& 0.631				& 0.556				\\ \hline
\end{tabular}
\end{small}
\caption{Object detection and semantic segmentation performance measured on the subset of Pascal VOC 2007 dataset.}
\label{table:det_seg}
\end{table}

\section{Conclusion}
\label{sec:conclusion}
In this work, we proposed a new image restoration method that is based on the estimation of the DCT coefficient distribution and showed that it can solve the JPEG artifact removal task well. 
The proposed architecture uses the typical encoder-decoder model in generating restored image by the help of the classifier output which is an estimated distribution of DCT coefficients of an image patch.
The resultant images generated by the proposed method have good visual quality with many sharp edges. Especially, our method is very good at removing the blocking artifacts and restoring high frequency texture information.  

{
\small
\bibliographystyle{ieee}
\bibliography{ref}
}

\end{document}